\documentclass[11pt,a4paper]{article}
\usepackage{times}
\usepackage{latexsym}
\usepackage{url}
\usepackage{graphicx}
\usepackage{algorithm2e}
\usepackage{amsmath, amssymb}
\usepackage{enumitem}
\usepackage{multirow}
\usepackage[hyperref]{naaclhlt2019}
\usepackage{comment}

\aclfinalcopy 

\title{A bag-of-concepts model improves relation extraction in a narrow knowledge domain with limited data}

\author{Jiyu Chen, Karin Verspoor \and Zenan Zhai  \\
School of Computing and Information Systems \\
The University of Melbourne, Australia \\
{\tt {\small jiyuc@student.unimelb.edu.au, \{karin.verspoor, zenan.zhai\}@unimelb.edu.au}}
}

\date{}

\begin{document}

\maketitle
\begin{abstract}

This paper focuses on a traditional relation extraction task in the context of limited annotated data and a narrow knowledge domain. We explore this task with a clinical corpus consisting of 200 breast cancer follow-up treatment letters in which 16 distinct types of relations are annotated. 
We experiment with  
an approach to extracting typed relations called window-bounded co-occurrence (WBC), which uses an adjustable context window around entity mentions of a relevant type,  
and compare its performance with a more typical intra-sentential co-occurrence baseline. We further introduce a new bag-of-concepts (BoC) approach to feature engineering based on the state-of-the-art word embeddings and word synonyms. We demonstrate the competitiveness of BoC by comparing with methods of higher complexity, and explore its effectiveness on this small dataset. 
\end{abstract}

\section{Introduction}
Applying automatic relation extraction on small data sets in a narrow knowledge domain is challenging. Here, we consider the specific context of a small clinical corpus, in which we have a variety of relation types of interest but limited examples of each. Transformation of clinical texts into structured sets of relations can facilitate the exploration of clinical research questions such as the potential risks of treatments for patients with certain characteristics, but large-scale annotation of these data sets is notoriously difficult due to the sensitivity of the data and the need for specialized clinical knowledge. 

Rule-based methods~\cite{abacha2011automatic,verspoor2016establishing} typically determine whether a particular type of relation exists in a given text 
by leveraging the context in which key clinical entities are mentioned. For instance, if two entities with type TestName and TestResult, respectively, are observed in a given sentence, it is likely that a relation of type TestFinding exists between them. However, construction of high-precision rules defining relevant contexts is time-consuming and expensive, requiring extensive effort from domain experts. 

The state-of-the-art machine learning algorithms such as neural network models~\cite{Nguyen:16,ammar2017ai2,DBLP:conf/emnlp/HuangW17} may over-fit in performing relation extraction in this context, due to a limited quantity of training instances.




In this work, we experiment with two automatic approaches to semantic relation extraction applied to a small corpus consisting of breast cancer follow-up treatment letters~\cite{pitson2017developing}, comparing a simple rule-based co-occurrence approach to machine learning classifiers.


The first approach, simple co-occurrence~\cite{verspoor2016establishing}, is based on the assumption that most relevant relations are intra-sentential, that is, the relation between a pair of named entities is expressed within the scope of a single sentence.
%
%
However, some relations may be expressed across sentence boundaries, and thus a single sentence may not be the 
ideal choice of scope, as shown in prior work that considers inter-sentential relations (also known as non-sentence or cross-sentence relations)~\cite{panyam2016asm,peng2017cross}.
We extend the co-occurrence approach to allow explicit adjustment of context window size, from one to two sentences, a method called  Window-Bounded Co-occurrence (WBC). 
The best window size for a given relation is identified by choosing the one which produces the highest score under F$_1$-measure on a development set.


The second approach is based on  supervised binary classification. We transform the multi-relation extraction task into several independent binary tasks. We build on a bag-of-concepts
(BoC)~\cite{sahlgren2004using} approach 
which models the text in terms of phrases or pre-identified concepts, 
 extending it with word embeddings and word synonyms. We compare two different pre-trained word embedding models, and a number of other model variations. We also explore grouping of synonyms into abstracted \textit{concepts}. 



We find that the intra-sentential rule-based approach 
outperforms the approach which allows for a larger context window. The supervised learning models outperforms rule-based approaches under F$_1$ measure, and their results show that models using BoC features outperform models with BoW, dependency parse, or sentence embedding features. We also show that SVM outperforms complex models such as a feed-forward ANN in our low resource scenario, with less tendency to over-fitting.

\section{Background}

At present, the two primary approaches to automatic relation extraction over biomedical corpora are rule-based approaches~\cite{verspoor2016establishing,abacha2011automatic} and machine learning approaches based on learners such as logistic regression, support vector machines (SVM)~\cite{panyam2016asm} and convolutional neural networks (CNN)~\cite{nguyen2018conv} together with sophisticated feature engineering methods. 

\citet{verspoor2016establishing} established a typical intra-sentential co-occurrence baseline with competitive performance comparing to a comprehensive machine learning-based system, PKDE4J~\cite{SONG2015320}, on the extraction of relations between human genetic variants and disease on the Variome corpus~\cite{verspoor2013annotating}. The sentential baseline is based on the assumption that the scope of relations is within one sentence, and further assumes that any pair of two entities mentioned in the same sentence and satisfying the type constraints of a given relation, expresses that relation. 
For example, if two entities with type TimeDescriptor and EndocrineTherapy respectively, the relation TherapyTiming will be extracted.
The sentential co-occurrence baseline set a benchmark for relation extraction on Variome corpus.

~\citet{abacha2011automatic} explored semantic rules for the extraction of relations between medical entities on PubMed Central (PMC) articles using linguistic patterns. They provide an example of implementing an end-to-end relation extraction system, applying named entity recognition in the first stage, then followed by the stage of relation extraction. They define several relation patterns based on medical knowledge, and leveraging the dependency parse tree of sentences in which entities occur. However,  the linguistic patterns and rules developed for their corpus likely are not directly applicable to our semantically distinct context of clinical letters.

 Machine learning methods vary based on the choice of models and the features considered. In model selection, the multi-type relation extraction task can be assigned to several independent binary  classifiers, each making the decision of whether a certain type of relation exist or not. 
 A basic binary classifier such as logistic regression
with ridge regularization is capable of performing relation extraction in this scenario. \citet{panyam2016asm} used support vector machines (SVM) with a dependency graph kernel to perform relation extraction on two biomedical relation extraction tasks, 
showing competitive results. Brown Clustering \cite{brown1992class} is a hierarchical approach to clustering words into classes through maximizing mutual information of bi-grams; it showed competitive performances in many NLP tasks \cite{turian2010word}.
\citet{nguyen2018conv} implemented a method using character-based word embeddings which can capture unknown words within the context, coupled with CNN and LSTM neural network models. This approach obtained  state-of-the-art performance in extracting chemical-disease relations on the BioCreative-V CDR corpus \cite{li2016biocreative}. 

For feature engineering, text features can generally be divided into the two categories of lexical features and syntactic features. Typical features used in other relation extraction tasks are summarized here.

\begin{itemize}
    \item \textbf{Bag-of-words (BoW) features} based on white-space delimited tokens, are used in many tasks as a starting point. 
    \item \textbf{Bag-of-concepts (BoC) features} ~\cite{sahlgren2004using} represent the text in terms of concepts, that is, phrases in the text that correspond to meaningful units. The current methods for generating concepts are based on techniques such as mutual information~\cite{sahlgren2004using}, or through dictionary-based strategies~\cite{funk2014bmcbionf}. For clinical texts, the  MetaMap tool~\cite{aronson2001effective} is often used to recognize clinical concepts.
    \item \textbf{Syntactic features}
    take sentence structure into account. For example, RelEx~\citep{fundel2006relex} uses dependency parse trees associated with small numbers of rules in extracting relations from MEDLINE abstract and reaches an overal 80\% precision and recall. Approximate Sub-graph Matching (ASM) \citep{liu2013approximate} enables sentences to be matched by considering the similarity of the structure of dependency parse subgraphs that connect relevant entities to subgraphs in the training data. 
    
    \item \textbf{Word embeddings} 
    aim to  
    capture word semantics through lower-dimension projections of word contexts and can be used to find word synonyms by measuring cosine similarity between word vectors. There are two widely used approaches to train word embeddings, co-occurrence matrix based methods such as GloVe~\cite{pennington2014glove}, and learning-based methods using skip-grams~\cite{mikolov2013skipgram} and CBOW~\cite{DBLP:journals/corr/abs-1301-3781}.
\end{itemize}

\section{Methods}
We improve the approaches described above to achieve better efficiency in relation extraction in our context of a narrow knowledge domain with limited data, specific to cancer follow-up treatment.
\subsection{Corpus}
We consider a previously introduced corpus related to breast cancer follow-up treatment, randomly sampled and manually annotated by two physicians \cite{pitson2017developing}. The corpus contains around 1000 sentences and 47,186 tokens. Despite its small size, the corpus is richly annotated with 16 medical named entity types and 16 types of semantic relations linking those entities with over 1,500 relation occurrences.
Entities within the corpus are related to clinical therapies, temporal events, diseases, and so on.  
The annotation of clinical relations includes the associations between identified entities. For example, in the context "She remains on Arimidex tablets.", "remains on" is a TimeDescriptor, and "Arimidex" is a EndocrineTherapy. The relation TherapyTiming holds between these two entities in this context (e.g.,  TherapyTiming(TimeDescriptor, EndocrineTherapy)). For conciseness, we use abbreviations to refer to the entity types; hence we will use TD to represent the entity type of TimeDescriptor. While the dataset hasn't been published yet, the full terminology list of entity types can be found in~\citet{pitson2017developing}.
To focus exclusively on the relation extraction task, we decouple the named entity recognition task from the relation extraction task by utilizing the gold standard entity annotations from the corpus.
\subsection{Method 1: Typed Sentential Co-occurrence}
\label{method1}
The simplest rule-based approach, given typed named entities, is to extract every pair of entities in a document that satisfies the type constraints of a relation. Such an approach  yields high  recall but poor precision, due to lack of  use of context needed to ensure that a specific relation is expressed as holding between the entities. 
For example, in our data, only the semantic relation of Toxicity is defined as connecting a Therapy to a ClinicalFinding (expressing that a therapy was found to cause a specific toxic effect) and so it might seem reasonable to assume that a Toxicity event is being expressed in a document where both a Therapy and a ClinicalFinding are mentioned. However, the occurrence of these two entity types together in a document do not strictly indicate an occurrence of Toxicity relation between them; the entities may be connected to other mentioned entities via different relations. Hence assuming a Toxicity relation between them would result in a false positive.

We used intra-sentential constraints as described by  \citep{verspoor2016establishing} to improve precision by only considering that named entities that co-occur in the same sentence can have valid semantic relations.
With this intra-sentential co-occurrence constraint, the relation extraction performance of the  sentential baseline achieved strong recall 
and competitive precision, as well as reasonable overall F-score,
 on the Variome corpus \cite{verspoor2013annotating}.



We introduce the approach of Window-Bounded Co-occurrence (WBC) to explore the impact of relaxing the constraint that the sentence boundary defines the scope of a relation. WBC defines a context window for a relation as the expansion to a base window of the sentence where an entity of the appropriate type occurs, and the adjustment of tokens beyond that sentence within which the related entity appears. We assume the occurrence of an inter-sentential relation relies on the distance of two entities, where  distance is defined based on the number of tokens considered beyond the base sentence in which an entity occurs. We introduce a hyperparameter, $\rho$, to represent the distance of an entity pair in a context, namely the number of tokens allowed in exceeding the single base sentence. $\rho=0$ denotes the entity pair is intra-sentential; $\rho=x$ denotes the second entity in a pair is $x$ number of tokens away from the base window. If $\rho$ is large than the length of the second context window, then the context window will set to the scope of two sentences by default.

\begin{figure}[hbtp]
\centering
\includegraphics[width=\columnwidth]{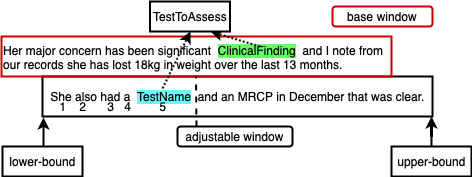}
\caption{Example of length-awareness sliding window of WBC in TestToAssess relation case}
\label{figure:sliding window}
\end{figure}

Figure~\ref{figure:sliding window} shows an example when $\rho = 5$, WBC allows for the extraction of the semantic relation of TestToAssess across the base window containing the entity of ClinicalFinding, and into the expanded window encompassing the subsequent sentence and containing the  TestName entity. 

\subsection{Method 2: Supervised Binary Classification Approach}
We adopt a traditional pipeline 
as the architecture of the relation extraction system. Each stage is introduced below.



\subsubsection{Data Preparation}
Considering the semantic variation in the texts, and the small number of examples, training several independent binary classifiers is more robust for mining individual type of semantic relation patterns.
Therefore, we transform the original dataset into 16 independent subsets,
by grouping instances by their relation type. An instance  consists of a typed entity pair, one or two sentence(s) with the relevant named entities inside as context, a label as an indication of relation occurrence, and a treatment letter id for mapping its position in the original dataset.
We remove four relation types with fewer than 10 annotated instances, specifically the relations Intervention, EffectOf, RecurLink and GetOpinion. 


We apply context selection for generating instances during data transformation. One instance represents the occurrence of a single semantic relation, containing one relevant entity pair. The context for each instance is the entire raw text of the sentence where the entity pair appears. In cases where more than one entity pair occurs in the same text context, we generate an independent instance for each entity pair. 
Where cross-sentence relations occur, we  concatenate the two sentences containing the relevant entities into a single sentence, structurally indicating that the two sentences are related. 
In each instance, we replace the two named entity phrases with their type. In the case of overlapping named entities, such as where one named entity partially or completely collides with another entity, both types are retained, adjacent to each other in the text. Other entities mentioned in the context not relevant for the specific relation are left in their original textual form.


We use NLTK \cite{bird2009natural} to perform tokenization and lemmatization to normalize the representation of the text, and strip punctuation. 
We use the Snowball English stopword list~\footnote{\url{http://anoncvs.postgresql.org/cvsweb.cgi/pgsql/src/backend/snowball/stopwords/}} 
 to remove stopwords.

Further details are presented in the feature engineering section below.

\subsubsection{Feature Engineering}
We implement a set of traditional semantic features and three main feature sets based on ASM, BoC, and sentence embeddings in the sections below.

\begin{itemize}[leftmargin=*]

\item \textbf{Traditional Semantic Features}\\
The traditional semantic features includes bag-of-words (count-based), lemmas (base, uninflected form of a noun or verb), algebraic expressions, named entity type (derived from the gold-standard), POS tags and dependency parse based on Stanford CoreNLP~\cite{manning2014stanford}, and a transformation from dependency parse tree to graph using NetworkX \cite{hagberg2008exploring} where edges are dependencies and nodes are tokens/labels. 

\item \textbf{ASM features}\\
The classical ASM measurement was developed by ~\citet{liu2013approximate}, and was later extended to kernel method by~\citet{panyam2016asm}. 
The ASM kernel was applied to the chemical induced disease (CID) task~\cite{wei2015overview} and Seedev shared  tasks~\cite{chaix2016overview}. The performance of ASM significantly depends on the result of POS-tagging and dependency tree parsing. All nodes are normalized to their lemmas. Here, Stanford CoreNLP~\cite{manning2014stanford} is used for POS tagging and dependency tree parsing of the text.
The context is split into sentences before dependency parsing is applied on individual sentences.

We produce the ASM features following \citet{panyam2016asm}. Where the context includes two sentences, a dummy root node is introduced to connect the root nodes of two dependency parse trees. Figure~\ref{figure:concatenation} shows an example. 
After pre-processing, the dependency tree structure is  transformed into a graph where nodes are lemmas with their POS tags, edges are dependencies across lemmas within sentence. Then, a shortest path algorithm is applied on the dependency graph to generate flat features.
%
In cases where sentences are very long, processing time is unacceptable, and no ASM features are generated.

\begin{figure}[ht]
\centering
\includegraphics[width=0.8\columnwidth]{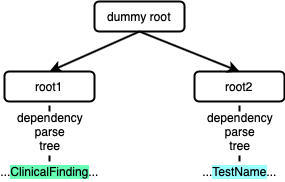}
\caption{Illustration of concatenation of two sentence parsing results using dummy root in TestToAssess relation setting}
\label{figure:concatenation}
\end{figure}

\item \textbf{Bag-of-Concepts features}\\
Word embeddings are used to capture word similarity based on shared surrounding context. The size of the surrounding context, known as window size control, varies the representation of word embeddings from more semantic (shorter window size) to more syntactic (longer window size). Synonyms can be identified by identifying two words with similar embeddings, based on cosine similarity measurement. 
  Over-fitting can occur for word embeddings, where a training corpus is not large enough or a corpus is limited to a narrow domain of knowledge. Therefore, instead of training word embeddings on our corpus, we use two publicly available pre-trained word embeddings, GloVe  \cite{pennington2014glove} and a Wikipedia-PubMed-PMC embedding~\cite{moen2013distributional} to capture more clinically relevant vocabulary.
The vocabulary of word embeddings denotes the total number of words that are represented. In our experiment, only the top 20,000 most frequent lemmas are selected. 
Gensim \cite{rehurek2010software} is used to find the synonyms of a lemma from the vocabulary by measuring similarity between GloVe word vectors.

We then implement an algorithm for building BoC. Using a word2concept algorithm (see  Equation~\ref{equation:w2c}), we map a lemma (key) to a concept (value) based on the embedding of the lemma expressed as $\pmb E(lemma)$ and the similarity threshold expressed as $\pmb \mu$ as a tunable hyper-parameter. 
\begin{equation}
CONCEPT = f(E(lemma), \mu)
\label{equation:w2c}
\end{equation}


The algorithm starts by extracting BoW features for each generated instance after data preparation process
into a list $L$. Then, starting from the first lemma $w_1$ from $L$, we retrieve its embedding $ x_{w_1} = E( w_1)$. We then retrieve a new lemma $w_i$ and its embedding $x_{w_i}$ from the vocabulary $V$, and calculate the similarity score $S = cos(x_{w_1}, x_{w_i})$. If $S >= \mu$, create mappings between $w_1\rightarrow concept_1$ and $w_i \rightarrow concept_1$. If no $w_i$ satisfies the condition of $S >= \mu$, then $w_1$ will be kept in its original form.   Next, we move to the second word $w_2$ in $L$, check whether $w_2$ has already been mapped to a concept $concept_*$, and if so, directly create the mapping $w_2 \rightarrow concept_*$. Otherwise, we iterate.

Note that in this model, named entities will effectively be treated as out-of-vocabulary terms, since they have been mapped to and replaced with the names of the relevant entity types (e.g., ``TestName'' which is not a token that would be expected to be represented in any pre-trained word embedding model).

\item \textbf{Sentence embedding features}\\
Apart from being a tool for finding word synonyms and generating BoC data representation, word embeddings can also be used to obtain sentence embeddings through weighted average pooling.
If $S$ denotes a sentence, $E(S)$ denotes the embedding of sentence $S$. $E(w_1)$ denotes the embedding of the first word $w_1$ of sentence $S$, $Score(w_1)$  denotes the TF-IDF score of word $w_1$, the sentence embeddings based on weighted average pooling can be expressed as Equation \ref{equation:word2sent}. We calculate TF-IDF scores of each word using the original documents, as each instance has an index to its original document id. 
Out-of-vocabulary tokens and gold standard named entities will be ignored. However, entity information has been considered during that data preparation stage, because the generation of instances takes gold standard entities into account.

\begin{equation}
E(S) = \frac{1}{n} \sum_{i=1}^{n}  E(w_i) \times Score(w_i)
\label{equation:word2sent}
\end{equation}

\end{itemize}


\subsection{Supervised Learning Models}
\begin{figure}[ht]
\centering
\includegraphics[width=0.8\columnwidth]{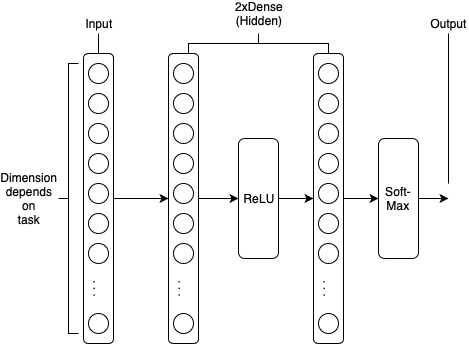}
\caption{Architecture of FNN}
\label{figure:fnn}
\end{figure}

We build individual binary classifiers for each relation type. We introduce the SVM classifiers and Feed-forward ANN models briefly here.
\begin{itemize}
    \item \textbf{Support Vector Machine} We select the general SVM model~\cite{Hearst:1998:SVM:630302.630387} and kernels provided by \texttt{scikit-learn}~\cite{scikit-learn}. For SVM kernels, we integrate ASM kernel as part of feature engineering, to avoid colliding with the use of the linear kernel and the RBF (Radial Basis Function) kernel.
    \item \textbf{Feed-forward ANN}
    We use Keras~\cite{chollet2015keras} 
    to construct a simple feed-forward neural network model with two fully connected layers as shown in Figure~\ref{figure:fnn}.
\end{itemize}


The dimension of each input and the number of hidden units in each layer is the same as the dimension of feature vectors under each type of relation. The activation function for the first dense layer is ReLU, and for the second dense layer is softmax.

\subsection{Experiment Design}
Considering the dataset is small, we split the transformed dataset into three independent combinations of training, development, and test sets with the ratio of 6:1:3. 
In training stage, 
we train independent models for each specific relation type. We use cross-validation and grid search to tune the hyperparameters of the classifiers.

In prediction stage,  the decision of applying sentence-bounded or window-bounded approach is made by setting the size of sliding window. Setting the window size to 0 will apply the sentence-bounded co-occurrence constraint. We choose window sizes of 0, 5, 10 to explore the value of additional context. 
In supervised binary classification approach, utilizing a similarity threshold of 1 leads to strict use of BoW (word) features, while relaxing the similarity threshold $\mu$ of 0.9, 0.8 will generate BoC (concepts). We compare the influence of different word embeddings in generating BoC based on their relation extraction performance on the test set.

Both rule-based approach and supervised binary classification approach will make predictions on the same test set, which allows empirical comparison between rule-based and machine learning approaches. 


We compare the impact of increasing data size for BoW and BoC by sub-sampling the training set into nine instance-incremental and non-overlapping sub-sets (combining them into sets representing 10\% to 90\% of the original training set) 
and visualize the performance variation. We explore whether word embeddings as a medium for generating BoC are more effective than the direct use of sentence embeddings
in cases where the dataset is small and knowledge domain is restricted to the specific domain of breast cancer treatment. We also explore the combination of BoC and sentence embeddings feature, in order to investigate how best to make integration of them 

Finally, 
we explore the possibility of applying more complex models for analysis of our small and specific knowledge domain dataset. We start from simple linear models including logistic regression and lin-SVM, then apply  rbf-SVM, feed-forward ANN on Keras with 32 batch per time, and epochs of 2, 10, 100, and 500 for each relation type.


\subsection{Evaluation}
We evaluate both overall and per relation type performance. Evaluation of the two approaches is performed on the same test set derived from the data preparation stage. In addition, considering the dataset is small, we calculate the mean score from three independent combinations of training, development and test sets. We evaluate results using micro F$_1$-measure since the number of positives and negatives were highly imbalanced across all relation types. We finally evaluate the performance growth over nine sub-sampled training sets of increasing size.



\section{Result and Discussion}
We present experiment results with micro F$_1$ measurement. After experimenting with different similarity threshold values to generate BoC features, the best performance is achieved when the threshold is set to 0.9 (results not shown). Word embeddings derived from Wiki-PubMed-PMC outperform GloVe-based embeddings (Table~\ref{table:results_per_feature}). The models using BoC outperform models using BoW as well as ASM  features.

As shown in Table \ref{table:results_per_type}, the intra-sentential co-occurrence baseline outperforms other approaches which allow boundary expansion. This is because a majority of relations in the corpus are intra-sentential.

\begin{figure}[!t]
\centering\includegraphics[width=.9\linewidth]{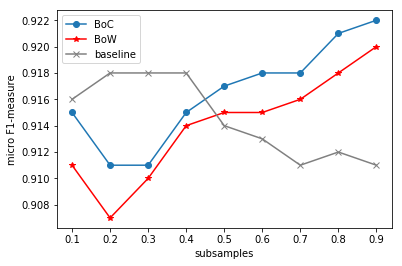}

\caption{Performance variation of BoW and BoC(\textsubscript{Wiki-PubMed-PMC, $\mu=0.9$}) with increasing data fractions, under linSVM, F$_1$ measure.}
\label{figure:f1growth}
\end{figure}

 A visualization of the growth in performance for both BoW and BoC-based models as training set size increases, over 12 relation types, based on  micro F$_1$, is shown in Figure \ref{figure:f1growth}. The results of BoC in this figure is collected from lin-SVM\textsubscript{Wiki-PubMed-PMC, $\mu=0.9$}, and BoW is collected from lin-SVM. We find BoC tends to outperform BoW with only a small number of training instances, and also performs better than BoW with incremental training instance.

\begin{table*}[!t]
    \centering
    \begin{tabular}{l|ccc|ccc|ccc}
         \hline
         Feature & \multicolumn{3}{c|}{LR} & \multicolumn{3}{c|}{SVM} & \multicolumn{3}{c}{ANN}\\
         \cline{2-10}
         & P & R & F$_1$ & P & R & F$_1$ & P & R & F$_1$ \\
         \hline
         +BoW & 0.93 & 0.91 & 0.92 & 0.94 & 0.92 & 0.93&0.91&0.91&0.91 \\
         +BoC (\textsubscript{Wiki-PubMed-PMC}) &0.94&0.92&\bf0.93& 0.94&0.92&\bf 0.93&0.91&0.91&\bf 0.91 \\
         +BoC (\textsubscript{GloVe}) &0.93&0.92&0.92&0.94&0.92&0.93&0.91&0.91&0.91 \\
         +ASM &0.90&0.85&0.88&0.90&0.86&0.88&0.89&0.89&0.89 \\
         +Sentence Embeddings(SEs) &0.89&0.89&0.89& 0.90&0.86&0.88&0.88&0.88&0.88\\
         +BoC(\textsubscript{Wiki-PubMed-PMC})+SEs&0.92&0.92&0.92&0.94&0.92&0.93&0.91&0.91&0.91\\
         \hline
    \end{tabular}
    \caption{Performance of supervised learning models with different features.}
    \label{table:results_per_feature}
\end{table*}

\begin{table*}[!t]
    \centering
    \begin{tabular}{l|c|ccc|ccc}
         \hline
         Relation type & Count & \multicolumn{3}{c|}{Intra-sentential co-occ.} & \multicolumn{3}{c}{BoC(\textsubscript{Wiki-PubMed-PMC})} \\
         \cline{3-8}
         & & $\rho=0$ & $\rho=5$ & $\rho=10$ & LR & SVM & ANN \\
         \hline
         TherapyTiming(\textsubscript{TP,TD}) &428&\bf 0.84&0.59&0.47&0.78&0.81&0.78 \\
         NextReview(\textsubscript{Followup,TP}) &164&\bf 0.90&0.83&0.63&0.86&0.88&0.84 \\
         Toxicity(\textsubscript{TP,CF/TR}) &163&\bf 0.91&0.77&0.55&0.85&0.86&0.86 \\
         TestTiming(\textsubscript{TN,TD/TP}) &184&0.90&0.81&0.42&0.96&\bf 0.97&0.95 \\
         TestFinding(\textsubscript{TN,TR}) &136&0.76&0.60&0.44&\bf 0.82&0.79&0.78 \\
         Threat(\textsubscript{O,CF/TR}) &32&0.85&0.69&0.54&\bf 0.95&\bf 0.95&0.92 \\
         Intervention(\textsubscript{TP,YR}) &5&\bf 0.88&0.65&0.47&-&-&- \\
         EffectOf(\textsubscript{Com,CF}) &3&\bf 0.92&0.62&0.23&-&-&- \\
         Severity(\textsubscript{CF,CS}) &75&\bf 0.61&0.53&0.47&0.52&0.55&0.51 \\
         RecurLink(\textsubscript{YR,YR/CF}) &7&\bf 1.0&\bf 1.0&0.64&-&-&- \\
         RecurInfer(\textsubscript{NR/YR,TR}) &51&0.97&0.69&0.43&\bf 0.99&\bf 0.99&0.98 \\
         GetOpinion(\textsubscript{Referral,CF/other}) &4&\bf 0.75&\bf 0.75&0.5& -&-&-\\
         Context(\textsubscript{Dis,DisCont}) &40&\bf 0.70&0.63&0.53&0.60&0.41&0.57\\
         TestToAssess(\textsubscript{TN,CF/TR})&36&0.76&0.66&0.36&\bf 0.92&\bf 0.92&0.91\\
         TimeStamp(\textsubscript{TD,TP})&221&\bf 0.88&0.83&0.50&0.86&0.85&0.83\\
         TimeLink(\textsubscript{TP,TP})&20&\bf 0.92&0.85&0.45&0.91&\bf 0.92&0.90\\
         \hline
         Overall &1569&0.90&0.73&0.45&0.92&\bf 0.93&0.91 \\
         \hline
    \end{tabular}
    \caption{ F$_1$ score results per relation type of the best performing models.}
    \label{table:results_per_type}
\end{table*}


The reason that Wikipedia-PubMed-PMC embeddings~\cite{moen2013distributional} outperforms GloVe~\cite{DBLP:journals/corr/abs-1301-3781} in the extraction of most relation types (Table \ref{table:results_per_feature}) is because its training corpus has a more similar domain and vocabulary as our dataset. Therefore, it leads to more relevant models of the distributional semantics of words.  On the other hand, the GloVe  embeddings are derived from a more general corpus; thus the semantics of domain specific terms in our dataset are not captured. 

By observation, most lemmas that map into concepts are digits, time stamps, common verbs and medical terminologies. For example, in TimeStamp(\textsubscript{TD,TP}), drugs with similar effects such as ``letrozole" and ``anastrozole" map into ``$CONCEPT_3$"; all single digits, ranges from 0-9 map into ``$CONCEPT_{26}$; year tags such as "2011, 2009" map into ``$CONCEPT_{56}$". We consider the cause of differences between the BoW and BoC representations. In the normalization process for BoW, stop-words collected from general knowledge domain are not well-suited to the knowledge domain for our specific task, while limited data does not allow the construction of an appropriate stop-words list from the data. In contrast, BoC models normalize the differences between individual words with shared meaning. While intuitively this should support an improvement over BoW models, we find BoW outperforms BoC when extracting certain relations such as TestTiming(\textsubscript{TN,TD}), TestToAssess(\textsubscript{TN,TR}) with gold standard named entities as arguments. The reason is that synonyms may express slightly different meanings;  concept mapping discards such  differences and leads to information loss, potentially causing more mis-classifications. The decision of whether to use the BoC or BoW will depend on the characteristics of particular relation types.

Table \ref{table:results_per_feature} also shows the combination feature of BoC and sentence embeddings outperforms sentence embeddings alone, but do not exceed the upper boundary of BoC feature, in which again demonstrating the competitiveness of BoC feature.

Since this corpus is much smaller than other narrow knowledge domain corpus such as CID ~\cite{wei2015overview} and Seedev shared task~\cite{chaix2016overview}, the training instances are not enough for the learners to generalize well using syntactic representation. Therefore, the models using ASM kernel~\cite{panyam2016asm} do not outperform the simple linear classifiers.

As the results of applying the co-occurrence baseline ($\rho=0$) shows (Table \ref{table:results_per_type}), the semantic relations in this data are strongly concentrated within a sentence boundary, especially for the relation of RecurLink, with an F$_1$ of 1.0. 
The machine learning approaches based on BoC lexical features effectively complement the deficiency of cross-sentence relation extraction.  

Lin-SVM outperforms other classifiers in extracting most relations.  The feed-forward ANN displays significant over-fitting across all relation types, as the performance decreases when increasing the training epochs. Specifically, with only two training epochs, the performance of ANN is still slightly worse than lin-SVM. The result of lin-SVM present its robustness of avoiding over-fitting compares to feed-forward ANN with BoW, BoC, ASM flat features and sentence embeddings. 




\section{Conclusion and Future Work}
We proposed two ways to perform relation extraction for a narrow knowledge domain,  with only small available data set. We implemented a window-based context approach and experiment with determining the best context size for the relation extraction in the rule-based settings. The typical sentential co-occurrence baseline is competitive when most relations are intra-sentential. 
We implemented a BoC feature engineering method, by leveraging word embeddings as a tool for finding word synonyms and mapping them to concepts. BoC feature outperforms BoW, ASM syntactic feature and sentence embeddings derived by weighted average pooling across word embeddings in small dataset with respect to its significant improvements in micro F$_1$ score. In addition, it would be expected to show competitive results on other relation extraction tasks where 
it is useful to generalize specific tokens such as digits or time stamps.

We also explored the performance of models with different level of complexity, such as logistic regression, lin-SVM, rbf-SVM, and a simple feed-forward ANN.  
The results highlight that strategies to avoid over-fitting must be considered since the number of training instances is limited.

In future work, we will explore a number of directions, including some unsupervised learning approaches. 
We will test the performance of BoC on other corpora, to explore BoC vs.\ BoW as a baseline data representation. 
We will address comparisons between BoC and other word clustering methods such as Brown Clustering \cite{brown1992class}.
Finally, the integration of current named entity recognition tools 
and end-to-end relation extraction, 
to remove the reliance on gold standard named entity annotations, will also be explored.

\bibliographystyle{acl_natbib}
\bibliography{naaclhlt2019}

\begin{thebibliography}{32}
\expandafter\ifx\csname natexlab\endcsname\relax\def\natexlab#1{#1}\fi

\bibitem[{Abacha and Zweigenbaum(2011)}]{abacha2011automatic}
Asma~Ben Abacha and Pierre Zweigenbaum. 2011.
\newblock {Automatic extraction of semantic relations between medical entities:
  a rule based approach}.
\newblock \emph{Journal of biomedical semantics}, 2(5):S4.

\bibitem[{Ammar et~al.(2017)Ammar, Peters, Bhagavatula, and
  Power}]{ammar2017ai2}
Waleed Ammar, Matthew Peters, Chandra Bhagavatula, and Russell Power. 2017.
\newblock {The AI2 system at SemEval-2017 Task 10 (ScienceIE): semi-supervised
  end-to-end entity and relation extraction}.
\newblock In \emph{Proceedings of the 11th International Workshop on Semantic
  Evaluation (SemEval-2017)}, pages 592--596.

\bibitem[{Aronson(2001)}]{aronson2001effective}
Alan~R Aronson. 2001.
\newblock {Effective mapping of biomedical text to the UMLS Metathesaurus: the
  MetaMap program.}
\newblock In \emph{Proceedings of the AMIA Symposium}, page~17. American
  Medical Informatics Association.

\bibitem[{Bird et~al.(2009)Bird, Klein, and Loper}]{bird2009natural}
Steven Bird, Ewan Klein, and Edward Loper. 2009.
\newblock \emph{{Natural language processing with Python: analyzing text with
  the natural language toolkit}}.
\newblock " O'Reilly Media, Inc.".

\bibitem[{Brown et~al.(1992)Brown, Desouza, Mercer, Pietra, and
  Lai}]{brown1992class}
Peter~F Brown, Peter~V Desouza, Robert~L Mercer, Vincent J~Della Pietra, and
  Jenifer~C Lai. 1992.
\newblock Class-based n-gram models of natural language.
\newblock \emph{Computational linguistics}, 18(4):467--479.

\bibitem[{Chaix et~al.(2016)Chaix, Dubreucq, Fatihi, Valsamou, Bossy, Ba,
  Del{\.e}ger, Zweigenbaum, Bessieres, Lepiniec et~al.}]{chaix2016overview}
Estelle Chaix, Bertrand Dubreucq, Abdelhak Fatihi, Dialekti Valsamou, Robert
  Bossy, Mouhamadou Ba, Louise Del{\.e}ger, Pierre Zweigenbaum, Philippe
  Bessieres, Loic Lepiniec, et~al. 2016.
\newblock {Overview of the Regulatory Network of Plant Seed Development
  (SeeDev) Task at the BioNLP Shared Task 2016.}
\newblock In \emph{Proceedings of the 4th BioNLP Shared Task Workshop}, pages
  1--11.

\bibitem[{Chollet et~al.(2015)}]{chollet2015keras}
Fran\c{c}ois Chollet et~al. 2015.
\newblock {Keras}.
\newblock \url{https://keras.io}.

\bibitem[{Fundel et~al.(2006)Fundel, K{\"u}ffner, and Zimmer}]{fundel2006relex}
Katrin Fundel, Robert K{\"u}ffner, and Ralf Zimmer. 2006.
\newblock {RelEx—Relation extraction using dependency parse trees}.
\newblock \emph{Bioinformatics}, 23(3):365--371.

\bibitem[{Funk et~al.(2014)Funk, Baumgartner, Garcia, Roeder, Bada, Cohen,
  Hunter, and Verspoor}]{funk2014bmcbionf}
Christopher Funk, William Baumgartner, Benjamin Garcia, Christophe Roeder,
  Michael Bada, K~Cohen, Lawrence Hunter, and Karin Verspoor. 2014.
\newblock {Large-scale biomedical concept recognition: an evaluation of current
  automatic annotators and their parameters}.
\newblock \emph{BMC Bioinformatics}, 15(1):59.

\bibitem[{Hagberg et~al.(2008)Hagberg, Swart, and
  S~Chult}]{hagberg2008exploring}
Aric Hagberg, Pieter Swart, and Daniel S~Chult. 2008.
\newblock {Exploring network structure, dynamics, and function using NetworkX}.
\newblock Technical report, Los Alamos National Lab.(LANL), Los Alamos, NM
  (United States).

\bibitem[{Hearst(1998)}]{Hearst:1998:SVM:630302.630387}
Marti~A. Hearst. 1998.
\newblock {Support Vector Machines}.
\newblock \emph{IEEE Intelligent Systems}, 13(4):18--28.

\bibitem[{Huang and Wang(2017)}]{DBLP:conf/emnlp/HuangW17}
Yi~Yao Huang and William~Yang Wang. 2017.
\newblock {Deep Residual Learning for Weakly-Supervised Relation Extraction}.
\newblock In \emph{Proceedings of the 2017 Conference on Empirical Methods in
  Natural Language Processing, {EMNLP} 2017, Copenhagen, Denmark, September
  9-11, 2017}, pages 1803--1807.

\bibitem[{Li et~al.(2016)Li, Sun, Johnson, Sciaky, Wei, Leaman, Davis,
  Mattingly, Wiegers, and Lu}]{li2016biocreative}
Jiao Li, Yueping Sun, Robin~J Johnson, Daniela Sciaky, Chih-Hsuan Wei, Robert
  Leaman, Allan~Peter Davis, Carolyn~J Mattingly, Thomas~C Wiegers, and Zhiyong
  Lu. 2016.
\newblock Biocreative v cdr task corpus: a resource for chemical disease
  relation extraction.
\newblock \emph{Database}, 2016.

\bibitem[{Liu et~al.(2013)Liu, Hunter, Ke{\v{s}}elj, and
  Verspoor}]{liu2013approximate}
Haibin Liu, Lawrence Hunter, Vlado Ke{\v{s}}elj, and Karin Verspoor. 2013.
\newblock {Approximate subgraph matching-based literature mining for biomedical
  events and relations}.
\newblock \emph{PloS one}, 8(4):e60954.

\bibitem[{Manning et~al.(2014)Manning, Surdeanu, Bauer, Finkel, Bethard, and
  McClosky}]{manning2014stanford}
Christopher Manning, Mihai Surdeanu, John Bauer, Jenny Finkel, Steven Bethard,
  and David McClosky. 2014.
\newblock {The Stanford CoreNLP natural language processing toolkit}.
\newblock In \emph{Proceedings of 52nd annual meeting of the association for
  computational linguistics: system demonstrations}, pages 55--60.

\bibitem[{Mikolov et~al.(2013{\natexlab{a}})Mikolov, Chen, Corrado, and
  Dean}]{DBLP:journals/corr/abs-1301-3781}
Tomas Mikolov, Kai Chen, Greg Corrado, and Jeffrey Dean. 2013{\natexlab{a}}.
\newblock \href {http://arxiv.org/abs/1301.3781} {Efficient estimation of word
  representations in vector space}.
\newblock \emph{CoRR}, abs/1301.3781.

\bibitem[{Mikolov et~al.(2013{\natexlab{b}})Mikolov, Sutskever, Chen, Corrado,
  and Dean}]{mikolov2013skipgram}
Tomas Mikolov, Ilya Sutskever, Kai Chen, Greg~S Corrado, and Jeff Dean.
  2013{\natexlab{b}}.
\newblock {Distributed representations of words and phrases and their
  compositionality}.
\newblock In \emph{Advances in neural information processing systems}, pages
  3111--3119.

\bibitem[{Moen and Ananiadou(2013)}]{moen2013distributional}
SPFGH Moen and Tapio Salakoski2~Sophia Ananiadou. 2013.
\newblock {Distributional semantics resources for biomedical text processing}.
\newblock In \emph{Proceedings of the 5th International Symposium on Languages
  in Biology and Medicine, Tokyo, Japan}, pages 39--43.

\bibitem[{Nguyen and Verspoor(2018)}]{nguyen2018conv}
Dat~Quoc Nguyen and Karin Verspoor. 2018.
\newblock {Convolutional neural networks for chemical-disease relation
  extraction are improved with character-based word embeddings}.
\newblock In \emph{Proceedings of the BioNLP 2018 workshop}, pages 129--136.
  Association for Computational Linguistics.

\bibitem[{Nguyen and Grishman(2016)}]{Nguyen:16}
Thien~Huu Nguyen and Ralph Grishman. 2016.
\newblock {Combining Neural Networks and Log-linear Models to Improve Relation
  Extraction}.
\newblock In \emph{Proceedings of IJCAI Workshop on Deep Learning for
  Artificial Intelligence (DLAI)}.

\bibitem[{Panyam et~al.(2016)Panyam, Verspoor, Cohn, and
  Kotagiri}]{panyam2016asm}
Nagesh~C Panyam, Karin Verspoor, Trevor Cohn, and Rao Kotagiri. 2016.
\newblock {ASM Kernel: Graph Kernel using Approximate Subgraph Matching for
  Relation Extraction}.
\newblock In \emph{Proceedings of the Australasian Language Technology
  Association Workshop 2016}, pages 65--73.

\bibitem[{Pedregosa et~al.(2011)Pedregosa, Varoquaux, Gramfort, Michel,
  Thirion, Grisel, Blondel, Prettenhofer, Weiss, Dubourg, Vanderplas, Passos,
  Cournapeau, Brucher, Perrot, and Duchesnay}]{scikit-learn}
F.~Pedregosa, G.~Varoquaux, A.~Gramfort, V.~Michel, B.~Thirion, O.~Grisel,
  M.~Blondel, P.~Prettenhofer, R.~Weiss, V.~Dubourg, J.~Vanderplas, A.~Passos,
  D.~Cournapeau, M.~Brucher, M.~Perrot, and E.~Duchesnay. 2011.
\newblock Scikit-learn: Machine learning in {P}ython.
\newblock \emph{Journal of Machine Learning Research}, 12:2825--2830.

\bibitem[{Peng et~al.(2017)Peng, Poon, Quirk, Toutanova, and
  Yih}]{peng2017cross}
Nanyun Peng, Hoifung Poon, Chris Quirk, Kristina Toutanova, and Wen-tau Yih.
  2017.
\newblock {Cross-Sentence N-ary Relation Extraction with Graph LSTMs}.
\newblock \emph{Transactions of the Association of Computational Linguistics},
  5(1):101--115.

\bibitem[{Pennington et~al.(2014)Pennington, Socher, and
  Manning}]{pennington2014glove}
Jeffrey Pennington, Richard Socher, and Christopher Manning. 2014.
\newblock {Glove: Global vectors for word representation}.
\newblock In \emph{Proceedings of the 2014 conference on empirical methods in
  natural language processing (EMNLP)}, pages 1532--1543.

\bibitem[{Pitson et~al.(2017)Pitson, Banks, Cavedon, and
  Verspoor}]{pitson2017developing}
Graham Pitson, Patricia Banks, Lawrence Cavedon, and Karin Verspoor. 2017.
\newblock {Developing a Manually Annotated Corpus of Clinical Letters for
  Breast Cancer Patients on Routine Follow-Up.}
\newblock \emph{Studies in health technology and informatics}, 235:196--200.

\bibitem[{Rehurek and Sojka(2010)}]{rehurek2010software}
Radim Rehurek and Petr Sojka. 2010.
\newblock {Software framework for topic modelling with large corpora}.
\newblock In \emph{In Proceedings of the LREC 2010 Workshop on New Challenges
  for NLP Frameworks}. Citeseer.

\bibitem[{Sahlgren and C{\"o}ster(2004)}]{sahlgren2004using}
Magnus Sahlgren and Rickard C{\"o}ster. 2004.
\newblock {Using bag-of-concepts to improve the performance of support vector
  machines in text categorization}.
\newblock In \emph{Proceedings of the 20th international conference on
  Computational Linguistics}, page 487. Association for Computational
  Linguistics.

\bibitem[{Song et~al.(2015)Song, Kim, Lee, Heo, and Kang}]{SONG2015320}
Min Song, Won~Chul Kim, Dahee Lee, Go~Eun Heo, and Keun~Young Kang. 2015.
\newblock {PKDE4J: Entity and relation extraction for public knowledge
  discovery}.
\newblock \emph{Journal of Biomedical Informatics}, 57:320 -- 332.

\bibitem[{Turian et~al.(2010)Turian, Ratinov, and Bengio}]{turian2010word}
Joseph Turian, Lev Ratinov, and Yoshua Bengio. 2010.
\newblock Word representations: a simple and general method for semi-supervised
  learning.
\newblock In \emph{Proceedings of the 48th annual meeting of the association
  for computational linguistics}, pages 384--394. Association for Computational
  Linguistics.

\bibitem[{Verspoor et~al.(2013)Verspoor, Jimeno~Yepes, Cavedon, McIntosh,
  Herten-Crabb, Thomas, and Plazzer}]{verspoor2013annotating}
Karin Verspoor, Antonio Jimeno~Yepes, Lawrence Cavedon, Tara McIntosh, Asha
  Herten-Crabb, Zo{\"e} Thomas, and John-Paul Plazzer. 2013.
\newblock {Annotating the biomedical literature for the human variome}.
\newblock \emph{Database}, 2013.

\bibitem[{Verspoor et~al.(2016)Verspoor, Heo, Kang, and
  Song}]{verspoor2016establishing}
Karin~M Verspoor, Go~Eun Heo, Keun~Young Kang, and Min Song. 2016.
\newblock {Establishing a baseline for literature mining human genetic variants
  and their relationships to disease cohorts}.
\newblock \emph{BMC medical informatics and decision making}, 16(1):68.

\bibitem[{Wei et~al.(2015)Wei, Peng, Leaman, Davis, Mattingly, Li, Wiegers, and
  Lu}]{wei2015overview}
Chih-Hsuan Wei, Yifan Peng, Robert Leaman, Allan~Peter Davis, Carolyn~J
  Mattingly, Jiao Li, Thomas~C Wiegers, and Zhiyong Lu. 2015.
\newblock {Overview of the BioCreative V chemical disease relation (CDR) task}.
\newblock In \emph{Proceedings of the fifth BioCreative challenge evaluation
  workshop}, pages 154--166.

\end{thebibliography}

\end{document}